\pdfoutput=1

\documentclass[11pt]{article}

\usepackage[]{ACL2023}
\usepackage{times}
\usepackage{latexsym}

\usepackage[T1]{fontenc}

\usepackage[utf8]{inputenc}

\usepackage{microtype}

\usepackage{inconsolata}
\usepackage{microtype}
\usepackage{enumitem}
\usepackage{placeins}
\usepackage{amssymb}
\usepackage{mathrsfs}
\usepackage{amsmath}
\usepackage{hhline}
\usepackage{amsfonts}
\usepackage{multirow}
\usepackage{array}
\usepackage{booktabs}
\usepackage{colortbl}
\usepackage{color}
\usepackage{subcaption}
\usepackage{threeparttable}
\usepackage{float}
\usepackage{caption}
\usepackage{graphicx}
\usepackage{arydshln}
\usepackage{makecell}
\usepackage{inconsolata}
\usepackage{framed} 
\usepackage{color}
\usepackage{svg}
\definecolor{shadecolor}{rgb}{0.92,0.92,0.92}
\title{ChatGPT vs Human-authored Text: Insights into Controllable Text Summarization and Sentence Style Transfer}
\author{Dongqi Liu \and Vera Demberg \\
        Department of Computer Science\\
        Department of Language Science and Technology\\
        Saarland Informatics Campus, Saarland University, Germany\\
        \texttt{\{dongqi,vera\}@lst.uni-saarland.de}}

\begin{document}
\maketitle
\begin{abstract}
Large-scale language models, like ChatGPT, have garnered significant media attention and stunned the public with their remarkable capacity for generating coherent text from short natural language prompts. In this paper, we aim to conduct a systematic inspection of ChatGPT's performance in two controllable generation tasks, with respect to ChatGPT's ability to adapt its output to different target audiences (expert vs.~layman) and writing styles (formal vs.~informal). Additionally, we evaluate the faithfulness of the generated text, and compare the model's performance with human-authored texts. Our findings indicate that the stylistic variations produced by humans are considerably larger than those demonstrated by ChatGPT, and the generated texts diverge from human samples in several characteristics, such as the distribution of word types. Moreover, we observe that ChatGPT sometimes incorporates factual errors or hallucinations when adapting the text to suit a specific style.\footnote{The project information of our study can be accessed at \href{https://dongqi.me/projects/ChatGPT_vs_Human}{https://dongqi.me/projects/ChatGPT\_vs\_Human}.}
\end{abstract}

\section{Introduction}

\textbf{\underline{G}}enerative \textbf{\underline{P}}re-trained \textbf{\underline{T}}ransformer (GPT; \textit{e.g.,} ChatGPT) models, which produce results from given conditional input prompts, have exhibited exceptional performance on various natural language understanding (NLU) and generation (NLG) tasks \cite{jiao2023chatgpt, wang2023cross, bang2023multitask, zhou2023comprehensive, dai2023chataug}. For instance, in NLU tasks, \citet{qin2023chatgpt} have proved that ChatGPT is comparable to state-of-the-art fine-tuning models in language reasoning. In NLG tasks, \citet{yang2023exploring} assessed four widely used benchmark datasets, such as QMSum, and confirmed ChatGPT's comparability to traditional fine-tuning methods. \citet{peng2023towards} further investigated effective strategies for machine translation using ChatGPT and highlight its strong translation ability. Additionally, ChatGPT can even facilitate multi-modal tasks \cite{yang2023mm, shen2023hugginggpt}, as well as the application of data augmentation \cite{dai2023chataug}. Although the studies mentioned above have demonstrated notable performance of ChatGPT across different domains, there remains a dearth of qualitative and quantitative evaluation of the texts generated by ChatGPT. Such an evaluation is vital to uncover the behavioral differences, potential limitations, and challenges associated with ChatGPT-generated texts, especially when compared with human-authored texts.

Controllable text generation seems to be a task in which ChatGPT-like models could potentially excel. This task is driven by the desire to tailor text for a diverse array of target users (\textit{e.g.,} experts and laypersons) \cite{kumar-etal-2022-gradient, cao-etal-2020-expertise, luo-etal-2022-readability}, and thereby enhancing the accessibility of textual information. In controllable text generation, one delineates a particular set of parameters or provides a prompt that defines the intended target style. This area has recently received growing interest from researchers in the field \cite{hu2021causal, li2022diffusion, zhang2022survey, dathathri2019plug, august-etal-2022-generating, carlsson-etal-2022-fine, gu-etal-2022-distributional, li2022diffusion, Keskar2019CTRLAC, Dathathri2019PlugAP}. The traditional natural language generation task \cite{pu-simaan-2022-passing}, which focuses solely on adequately responding with respect to a given input, can be regarded as a special case of controllable natural language generation, wherein the control setting remains unconditioned. Considering ChatGPT as the most recent language generation capability, the assessment of its language generation proficiency, specifically in the realm of controllable language generation, remains largely uncharted. Therefore, our study delves into two distinct applications of ChatGPT, namely controllable summary generation and sentence style transfer. In the former, we examine ChatGPT's ability to generate summaries that cater to two distinct readerships, namely experts and non-experts, for a given academic literature. Concerning sentence style transfer, we investigate ChatGPT's capability to generate both formal and informal sentences for the task of sentence formality.

The objective of this study is to tackle the research question: \textbf{In relation to the human-produced text, to what extent does ChatGPT-generated content demonstrate significant divergence from human behavior and the potential susceptibility to inaccuracies?} Our primary contributions are enumerated below:

\begin{itemize}
[itemsep=1pt,topsep=1pt,parsep=1pt]
    \item To the best of our knowledge, we are the first to utilize ChatGPT to evaluate its effectiveness in controllable text generation. 
    \item Our findings indicate that there are substantial performance disparities between the text generated by ChatGPT and that generated by humans.
    \item Our study exposes and quantifies the existence of numerous hard-to-spot errors in the text generated by ChatGPT, which have a tendency to amplify with successive transformations of the text.
\end{itemize}

\section{Related Work}
\subsection{Controllable Text Summarization}

Controllable text summarization is a rapidly evolving field that aims to produce summaries with specific characteristics, such as length, style, or content \cite{shen-etal-2022-mred, chan-etal-2021-controllable, sarkhel-etal-2020-interpretable, shen-etal-2022-sentbs, goldsack-etal-2022-making, Keskar2019CTRLAC, Dathathri2019PlugAP, he-etal-2022-ctrlsum, earle2021learning, liu-etal-2022-length}. A range of approaches has been proposed for this task, including the use of sequence-to-sequence models such as the Transformer model \cite{vaswani2017attention}. These models have demonstrated promising progress in producing high-quality summaries that can be modulated according to specific requirements \cite{fan-etal-2018-controllable, wu-etal-2021-controllable, amplayo-etal-2021-aspect}. Additionally, other techniques also have been proposed to enhance the controllability of the summaries, such as conditional generation \cite{he-etal-2022-ctrlsum, luo-etal-2022-readability}, prompt-based summarization \cite{yang2022tailor, liu-etal-2022-psp, zhang-song-2022-discup}, and multi-task learning \cite{cui-hu-2021-topic-guided, gu-etal-2022-distributional}.

\subsection{Text Style Transfer}

Text style transfer is a task that involves transforming an input sentence into a desired style while retaining its style-independent semantics \cite{jin-etal-2022-deep, zhu-etal-2021-multimodal, dai-etal-2019-style, li-etal-2020-dgst, babakov-etal-2022-large, mir-etal-2019-evaluating, ramesh-kashyap-etal-2022-different, tokpo-calders-2022-text}. To achieve this, prior research has examined sequence-to-sequence learning strategies that utilize parallel corpora with paired source/target sentences in different styles \cite{cheng-etal-2020-contextual, hu-etal-2021-syntax, nouri-2022-text}. Owing to the considerable demand for human resources and material investments in data labeling, parallel data across diverse styles are scarce. This has led to an increased interest in exploring more pragmatic situations where only non-parallel stylized corpora are accessible \cite{malmi-etal-2020-unsupervised, reif-etal-2022-recipe}.

\subsection{ChatGPT}
ChatGPT\footnote{https://openai.com/blog/chatgpt} is a large language model (LLM), which is built upon the innovations and improvements of its predecessors, such as GPT-3\footnote{https://openai.com/research/instruction-following}. In terms of training strategies, ChatGPT employs instruction learning and reinforcement learning from human feedback \cite[RLHF; ][]{ouyang2022training} to enhance its overall performance and adaptability.

Upon its emergence, ChatGPT has garnered considerable attention from researchers, who have undertaken initial studies into the model. Scholars such as \citet{baidoo2023education, rudolph2023chatgpt, west2023ai, sobania2023analysis, gilson2023does, lai2023chatgpt, wang2023zeroshot} have explored the notable strengths of ChatGPT from the fields of education, science, programming, healthcare, and text generation, respectively. However, \citet{Bang2023AMM} discovered that ChatGPT suffers from hallucination issues in the context of logical reasoning. Due to its immense and inaccessible training corpus and parameters, and the inability to access external knowledge for reliable sources of support, it is imperative to question whether ChatGPT demonstrates the same hallucination issue as other LLMs when performing sentence generation. Based on these clues, we firmly assert that in-depth analysis of the text generated by ChatGPT and its behavioral patterns are both significant and valuable, and can provide meaningful insights to the readers of this paper.

\section{Study on Controllable Summarization}

\subsection{Prompt Formulation}
In this section, our main objective is to test the zero-shot performance of ChatGPT on controllable summarization, with the goal to generate summaries for laymen vs. experts. To this end, we constructed several prompts as natural language instructions for ChatGPT. The prompts we tested include for the layman style: \textit{Please give me a layman / simple / simplified and understandable / easy-to-comprehend / straightforward / general audience \underline{summary} of X}, where $X$ was replaced by the source text that should be summarized. Similarly, for the expert summary, we experimented with the prompts: \textit{Please give me an expert / a technical / comprehensive and detailed / difficult-to-comprehend / in-depth / complicated \underline{summary} of X}.

\subsection{Experimental Setup}
For all experiments, we used ChatGPT \textit{gpt-3.5-turbo}, which was, at the time of experimentation, the best-performing publicly accessible version provided by OpenAI. For the hyper-parameter setting, we set temperature = 0, top p = 1, frequency penalty = 0.2, and presence penalty = 0.2. For summary generation, we configured the maximum number of generated tokens to 512. The remaining hyper-parameters were set to their default values as recommended by OpenAI. It is noteworthy that ChatGPT has the potential to generate empty responses (i.e., empty strings) as the result of network transmission timeouts or API request overloads. Should this arise, we adhere to the established practice of resubmitting the request until ChatGPT provides non-empty responses.

All of our experiments were conducted on the version of ChatGPT between 15 Feb 2023 and 30 Apr 2023 by using the OpenAI’s ChatGPT API.\footnote{https://platform.openai.com/overview} We should emphasize that to prevent any potential interference from the prior responses, we cleared the conversation history each time we submit a new query to ChatGPT. Unless otherwise specified, we refrained from engaging in any further conversation with ChatGPT to modify its responses.

\subsection{Dataset}

We selected ELIFE \cite{goldsack-etal-2022-making} dataset for our experiments. It contains summaries of academic literature that exhibit varying levels of readability, tailored to suit either expert or non-expert audiences. By means of this dataset, we can examine to what extent ChatGPT can regulate the summary generation process in accordance with the intended target users, and compare its summaries to human summaries. 

\subsection{Metrics}
In order to assess automatically whether ChatGPT summaries substantially differ in terms of their audience design based on the given prompt, we opted for a set of three automatic readability metrics: Flesch Reading Ease \cite[FRE; ][]{kincaid1975derivation}, Coleman-Liau Index \cite[CLI; ][]{coleman1975computer}, and Dale-Chall Readability Score \cite[DCR; ][]{chall1995readability}. 

The Flesch Reading Ease \cite{kincaid1975derivation} is a metric that gauges the comprehensibility of a given text. This index relies on the average number of syllables per word and the average number of words per sentence. A higher score signifies an easier-to-understand text. Additionally, the Coleman-Liau Index \cite{coleman1975computer} is a measure of the text's difficulty level, which considers the average number of characters per sentence and the average number of sentences per 100 words. A higher score indicates a more challenging text. The Dale-Chall Readability Score \cite{chall1995readability} is computed by comparing the number of complex words in the text with a list of common words. A higher score denotes a more challenging text. 

We also employed Rouge scores \cite{lin-2004-rouge} to evaluate the performance of ChatGPT in the task of text summarization, with the aim of comparing its efficacy against the state-of-the-art model. In order to assess the extent to which the summaries re-use word sequences from the original text, we furthermore evaluated N-gram novelty \cite{see-etal-2017-get, gehrmann-etal-2019-generating, pu-etal-2022-two}. Finally, we quantified inconsistency based on factual consistency checking metric SummaC \cite{laban-etal-2022-summac}, as well as hallucination checking metric \cite{cao-etal-2022-hallucinated, fischer2021finding}. SummaC \cite{laban-etal-2022-summac} uses sentence compression and summarization techniques to extract important information and improve the detection of inconsistencies in NLI models by segmenting documents and aggregating scores. Named entity hallucination \cite{fischer2021finding} flags potential hallucinations in named entities if they do not match the original sources. We here used BERT semantic similarity, rather than exact matching, when computing the named entities matching.

\subsection{Results on Controllable Summarization}

\subsubsection{Effect of Prompt Formulation}
\label{3.5.1}
Table \ref{tab: Prompts Reading Difficulty - elife} illustrates that different prompt versions are somewhat consistent regarding whether the instructions asking for layman summaries actually lead to more readable texts than those asking for expert summaries, with FRE ranging between scores of 31 and 38 for automatically generated layman summaries, and between 28 and 37 for automatically generated expert summaries. Conversely, human-written summaries exhibit very large differences according to the automatic metrics, with FRE of 53.1 for layman summaries and 22.5 for expert summaries. Similar effects are observed for the CLI and DCR measures. This preliminary test was conducted on a subset of the ELIFE dataset, containing merely 500 random samples; for the rest of the tests, we proceeded to the entire dataset, selecting the prompts asking for ``layman'' and ``expert'' summaries, as responses for these prompts seemed to align with the right direction wrt.~the readability measures.

\begin{table}[htbp]
\centering
\scalebox{0.9}{\tabcolsep=4pt
\begin{threeparttable}
\begin{tabular}{l c c c}
\toprule
Prompt version & FRE & CLI & DCR \\
\hline
layman & 37.26$^\dag$ & 14.82$^\dag$ & 11.21$^\dag$ \\
{simple}  & 31.92$^\dag$ & 15.70$^\dag$ & 11.54$^\dag$ \\
{simplified and understand.}  & 35.48$^\dag$ & 15.17$^\dag$ & 11.21$^\dag$ \\
{easy-to-comprehend}  & 36.59$^\dag$ & 14.93$^\dag$ & 11.32$^\dag$ \\
{straightforward}  & 31.74$^\dag$ & 15.58$^\dag$ & 11.42$^\dag$ \\
{general audience}  & 35.86$^\dag$ & 14.98$^\dag$ & 10.96$^\dag$ \\
\hdashline
human answer (for layman) & 53.06 & 12.36 & 8.90\\
\hline
expert & 29.89$^\dag$ & 15.91$^\dag$ & 11.88$^\dag$\\
{technical}  & 36.65$^\dag$ & 13.76$^\dag$ & 12.20$^\dag$\\
{comprehensive and detailed}  & 31.62$^\dag$ & 15.47$^\dag$ & 11.15$^\dag$ \\
{difficult-to-comprehend}  & 28.95$^\dag$ & 16.14$^\dag$ & 11.71$^\dag$ \\
{in-depth}  & 34.37$^\dag$ & 14.93$^\dag$ & 10.82$^\dag$ \\
{complicated}  & 29.05$^\dag$ & 15.76$^\dag$ & 11.40$^\dag$ \\
\hdashline
human answer (for expert) & 22.54 & 17.65 & 11.79 \\
\bottomrule
\end{tabular}
\end{threeparttable}
}
\caption{Reading difficulty on different prompts, tested on a set of 500 randomly selected items from the dataset. $^\dag$ indicates statistical significance (p$<$0.05) against corresponding human answers via paired t-test.}
\label{tab: Prompts Reading Difficulty - elife}
\end{table}

\subsubsection{Reading Difficulty Control}

Table \ref{tab: Reading Difficulty} corroborates that the results of the whole dataset are consistent with the findings from the smaller sample. We conclude that ChatGPT can produce summaries with different levels of reading difficulty to a certain extent based on the provided prompts. Notably, ChatGPT-generated sentences for expert-style summaries show greater complexity than those for layman-style summaries. However, the magnitude of the difference in the reading difficulty scores between the two types of summaries is considerably smaller than that observed in human-written summaries.

\begin{table}[ht]
\centering
\scalebox{1}{\tabcolsep=4pt
\begin{threeparttable}
\begin{tabular}{c c c c}
\toprule
Candidate & FRE & CLI & DCR \\
\hline
Human Layman & 52.42 & 12.46 & ~8.93 \\
Human Expert & 23.20 & 17.62 & 11.78 \\
ChatGPT Layman & 37.38$^\dag$$^\ddag$ & 14.78$^\dag$$^\ddag$ & 11.17$^\dag$$^\ddag$ \\
ChatGPT Expert & 30.38$^\dag$$^\ddag$ & 15.82$^\dag$$^\ddag$ & 11.85$^\dag$$^\ddag$ \\
\bottomrule
\end{tabular}
\end{threeparttable}
}
\caption{Reading difficulty scores by automatic metrics; 
$^\dag$ and $^\ddag$ indicate statistical significance (p$<$0.05) against same-style human answers, and opposite-style ChatGPT answers via paired t-test, respectively.}
\label{tab: Reading Difficulty}
\end{table}

\subsubsection{Comparison to Previous SOTA Model}
We also compared summaries generated by ChatGPT to a previous state-of-the-art (SOTA) neural fine-tuned summarization model \cite{pu2023incorporating}. On the same test split, the summaries produced by ChatGPT reached Rouge-1=25.53, Rouge-2=5.48, Rouge-L=13.30 under unsupervised learning, and Rouge-1=47.88, Rouge-2=13.75, Rouge-L=42.44 in few-shot learning use the training samples from the same subset of Section \ref{3.5.1}, while the model by \citet{pu2023incorporating} reached Rouge-1=48.70, Rouge-2=14.84, and Rouge-L=46.13.

\subsubsection{Disparities in Summarization Behavior}
We next examined whether ChatGPT and Humans are consistent with each other regarding the readability of summarization with respect to different items -- it could be possible, that some texts simply lead to less readable summaries than others. However, we discovered that Pearson correlations of FRE scores for summaries by humans and ChatGPT were only 0.31 for expert summaries, and 0.2 for layman summaries. (Scores were similarly low for the CLI and DCR metrics.) In addition, the statistical significance test elucidates the noteworthy divergence between the distinctive response styles produced by ChatGPT and the analogous styles of human-generated answers.

\begin{figure}[h]
  \centering
  \includegraphics[width=1\columnwidth,height=0.3\textwidth]{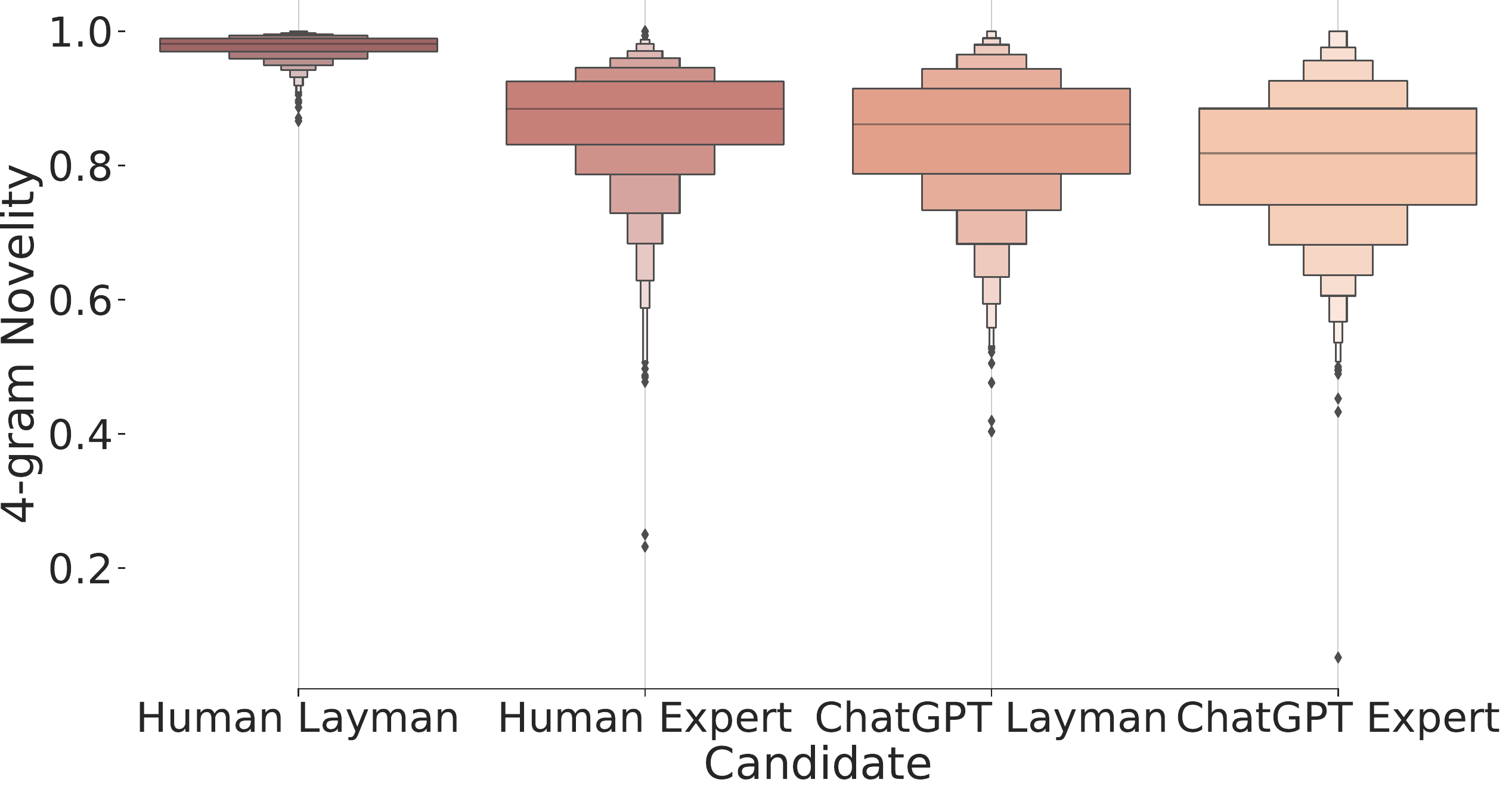}
  \caption{Comparison of abstractiveness between ChatGPT and human-generated summaries}
  \label{fig:n_gram_novelty}
\end{figure}

Following this, we contrasted the n-gram novelty of human vs.~ChatGPT summaries wrt.~the original texts. Figure \ref{fig:n_gram_novelty} reveals that a significantly higher number of novel 4-grams are present in human-written summaries, particularly those aimed at laymen. This suggests that ChatGPT summaries are slightly more extractive compared to human summaries.

\subsubsection{Inconsistencies and Hallucinations}
\begin{figure}[ht]
  \centering
  \includegraphics[width=0.5\textwidth,height=0.35\textwidth]{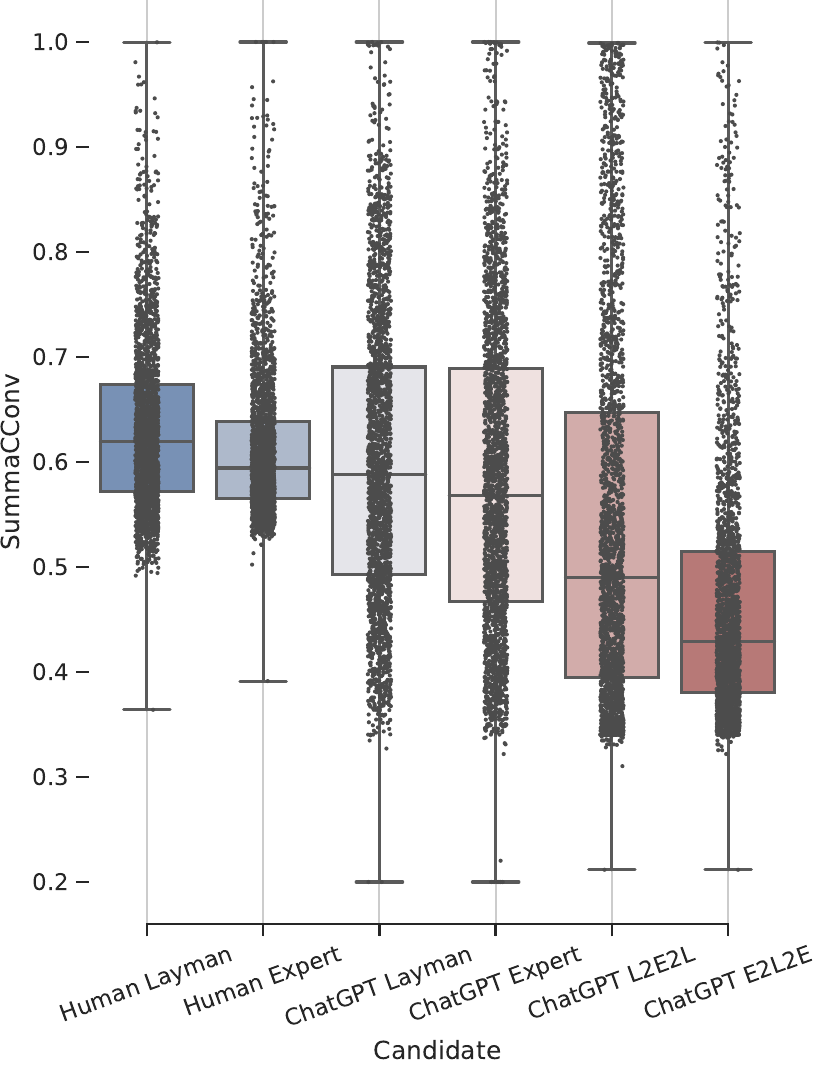}
  \caption{Summary consistency detection. L stands for layman, E for expert.}
  \label{fig:SUMMAC}
\end{figure}

Given that ChatGPT has previously been reported to generate misinformation, we sought to evaluate its risk of hallucinating on our specific task. Figure \ref{fig:SUMMAC} demonstrates that the SummaC consistency scores are lower for ChatGPT-generated summaries than for human-written summaries. A corresponding phenomenon is verified in the hallucination assessment. Precision scores provided in Table \ref{tab: Named Entity Hallucination - Elife} demonstrates the extent to which ChatGPT-generated text contains named entities that are absent in the source text. A lower precision score suggests that the generated text has more named entities that lack support in the source text. The recall scores reflect the ability of ChatGPT to capture named entities from the source text. A lower recall score implies that ChatGPT has missed a considerable number of named entities from the source text. F1 score represents the harmonic mean of the precision and recall scores. By examining Table \ref{tab: Named Entity Hallucination - Elife}, our findings demonstrate that ChatGPT generates a greater number of named entities that are not present in the source text after undergoing multiple iterations of text conversions and modification. For example, in an expert summary, ChatGPT misinterpreted the meaning of ``Geocode'' as ``regional regulations''.

\begin{table}[htbp]
\centering
\scalebox{1}{\tabcolsep=4pt
\begin{threeparttable}
\begin{tabular}{c c c c}
\toprule
Candidate & Precision & Recall & F1 \\
\hline
Human Layman & 0.78 & 0.63 & 0.70\\
Human Expert & 0.92 & 0.61 & 0.73\\
ChatGPT Layman & 0.75$^\ddag$ & 0.47$^\dag$ & 0.58$^\dag$ \\
ChatGPT Expert & 0.90$^\ddag$ & 0.49$^\dag$ & 0.63$^\dag$ \\
ChatGPT L2E2L & 0.74$^\ddag$ & 0.39$^\dag$$^\ddag$ & 0.51$^\dag$$^\ddag$ \\
ChatGPT E2L2E & 0.88$^\ddag$ & 0.47$^\dag$$^\ddag$ & 0.62$^\dag$$^\ddag$ \\
\bottomrule
\end{tabular}
\end{threeparttable}
}
\caption{Named entity hallucination on Elife dataset. $^\dag$ and $^\ddag$ indicate statistical significance (p$<$0.05) against same-style human answers, and opposite-style ChatGPT answers via paired t-test, respectively. L stands for layman, E for expert.}
\label{tab: Named Entity Hallucination - Elife}
\end{table}

\subsection{Intermediary Discussion}
Our experiments show that ChatGPT-generated summaries do not adapt as strongly to the target audience as human-authored summaries. One possible reason could be that ChatGPT, given the zero-shot setting, had no way to ``know'' how strongly the texts should be adapted to the target style. Furthermore, we identified evidence for potential hallucinations generated during summarization. We, therefore, carried out two post-hoc experiments: (1) We modified the prompt to include an example from the dataset, so ChatGPT would have a chance to know the expected level of text adaptation. (2) We subjected the resulting summaries to several re-writing steps and test whether this further intensifies the occurrence of hallucinations.

\subsubsection{Follow-up Experiment: Example Inclusion in Prompt}
We experimented with prompts that also include a human summary example. Unlike the previous few-shot learning experiment, we do not adjust the parameters of the ChatGPT, but just let the model perform unsupervised reasoning through the contents of the prompt. We observe (see Appendix Table \ref{tab:Reading Difficulty of One-shot Guidance}) that when guided by a human example from the dataset, the summaries generated by ChatGPT indeed tend to be more aligned with human performance, particularly on the Flesch Reading Ease metric (49.23 for layman, 28.88 for expert summaries). However, no significant changes are detected in other metrics. The degree of control over the summarization style has increased, yet it remains inferior to human capabilities. 

\subsubsection{Follow-up Experiment: Repeated Re-writing}
\label{3.8}
Summaries are further re-written based on the prompt \textit{Please give me a \textbf{layman}/\textbf{expert} style \underline{version} of $X$}, where $X$ was the previously generated summary. Figure \ref{fig:SUMMAC} and Table \ref{tab: Named Entity Hallucination - Elife} display the performance of ChatGPT after re-writing in the entries ``ChatGPT L2E2L" and ``ChatGPT E2L2E'' which stand for the order in which instructions were given (L stands for layman, and E for expert). The examinations point out that misinformation and hallucinations may be further increased during subsequent rewriting (lower SummaC scores, lower values in the named entity hallucination metric).

\section{Study on Text Formality Transfer}

\subsection{Prompt Formulation and Experimental Setup}
Our subsequent set of experiments investigates ChatGPT's capacity for style transfer concerning language formality. Our prompt for this task was formulated as \textit{Please give me a \textbf{formal} / an \textbf{informal} \underline{version} of $X$}. We utilized the same experimental setup as for the summarization task; however, we restricted the maximum number of generated tokens to 32. We again experimented with various prompts, as shown in Table \ref{tab: Prompts Reading Difficulty - GYAFC} below. Unless otherwise specified, all experiments used the same configuration.

\subsection{Dataset}
We investigated whether ChatGPT can proficiently execute style transfer on sentences using data from the GYAFC \cite{rao-tetreault-2018-dear} dataset. The dataset has two branches, Entertainment \& Music (EM) and Family \& Relationships (FR). With the aid of this dataset, we aim to evaluate ChatGPT's ability for sentence style transfer, examine the differences in vocabulary selection and syntactic structures between ChatGPT and human performance, and identify the limitations of ChatGPT.

\subsection{Metrics}
To evaluate the level of formality in the generated text, we utilized Text Formality Score \cite{heylighen1999formality} and MTLD Lexical Diversity \cite{mccarthy2010mtld} metric. The Text Formality Score \cite{heylighen1999formality} is a metric that quantifies the degree of formality in language usage within a text, based on the adherence to formal linguistic norms. Another measure that evaluates language formality is the MTLD Lexical Diversity metric \cite{mccarthy2010mtld}. This index measures the diversity and richness of the vocabulary used in the text, based on the frequency and number of unique words. A higher MTLD score indicates a greater variety of vocabulary, which typically corresponds to a more formal language style. We also utilized BLEU \cite{papineni-etal-2002-bleu} score to draw a comparison between ChatGPT and SOTA approach. We additionally assessed the distribution of POS tags in the generated different styles, as well as the distribution of dependency labels\footnote{https://spacy.io/}. For quantifying misinformation and hallucinations, we used DAE and named entity hallucination checking. The DAE algorithm \cite{goyal-durrett-2020-evaluating} utilizes dependency arcs to identify entailment relationships between propositions and identify inconsistencies in factual information based on syntactic and semantic structures.

\subsection{Results on Formality Control} 
\subsubsection{Effect of Prompt Formulation}
\label{4.4.1}

Table \ref{tab: Prompts Reading Difficulty - GYAFC} presents the results for a set of 500 random samples from the GYAFC dataset. We observe that the Formality scores are very similar for ChatGPT formal vs.~informal texts. We note however that the difference in ratings for human-written texts is also small for this metric. The MTLD metric on the other hand shows higher values for ChatGPT-generated formal texts; in fact, the scores are substantially larger than those of human-written texts, but differ not much from each other. We therefore proceed with the prompts using the formulation formal/informal for the rest of the experiments on the whole dataset. 

\begin{table}[htbp]
\centering
\scalebox{0.95}{\tabcolsep=4pt
\begin{threeparttable}
\begin{tabular}{l c c}
\toprule
Prompt version & Formality & MTLD \\
\hline
informal & 51.09 & 13.22$^\dag$ \\
{unprofessional} & 51.20 & 16.23$^\dag$ \\
{spoken version}  & 51.30$^\dag$ & 14.47$^\dag$ \\
{easygoing}  & 51.43$^\dag$ & 14.11$^\dag$ \\
{casual}  & 51.00 & 16.30$^\dag$ \\
{laid-back}  & 51.27 & 13.94$^\dag$ \\
\hdashline
human answer (for informal) & 50.76 & 11.42 \\
\hline
formal & 52.22$^\dag$ & 31.23$^\dag$ \\
{professional}  & 51.96$^\dag$ & 31.98$^\dag$ \\
{written}  & 51.62$^\dag$ & 29.69$^\dag$ \\
{stately}  & 51.30$^\dag$ & 34.43$^\dag$ \\
{grandiose}  & 52.85$^\dag$ & 30.71$^\dag$ \\
{majestic}  & 52.23$^\dag$ & 33.49$^\dag$ \\
\hdashline
human answer (for formal) & 53.92 & 14.99 \\
\bottomrule
\end{tabular}
\end{threeparttable}
}
\caption{Text formality on different prompts, tested on a set of 500 randomly selected items from the dataset. $^\dag$ indicates statistical significance (p$<$0.05) against corresponding human answers via paired t-test.}
\label{tab: Prompts Reading Difficulty - GYAFC}
\end{table}

\subsubsection{Sentence Formality Control}

Table \ref{tab: Text Formality} offers supplementary evidence from the full dataset supporting ChatGPT's capacity to modify the formality level of sentences. By employing the Formality indicator \cite{heylighen1999formality}, it is apparent that the generated text tends to manifest a higher level of formality overall. A primary factor contributing to this result is the predisposition of ChatGPT's training corpus towards written sources, encompassing materials such as books and news articles, as opposed to spoken language corpora \cite{OpenAI2023GPT4TR}. This perspective is further corroborated by an examination of the generated sentence samples. The MTLD metric underscores that ChatGPT's lexical diversity is considerably lower when generating informal sentences, but shows a marked increase when generating formal sentences. 

\begin{table}[htb]
\centering
\scalebox{1}{\tabcolsep=4pt
\begin{threeparttable}
\begin{tabular}{c c c c}
\toprule
Dataset & Candidate & Formality & MTLD \\

\hline
\multirow{4}*{\rotatebox{90}{\small GYAFC-FR}}
& Human Informal & 49.87 & 15.20 \\
~ & Human Formal & 53.57 & 18.70 \\
~ & ChatGPT Informal & 50.77$^\dag$$^\ddag$ & 14.60$^\ddag$ \\
~ & ChatGPT Formal & 52.06$^\dag$$^\ddag$ & 31.68$^\dag$$^\ddag$ \\
\midrule
\midrule

\multirow{4}*{\rotatebox{90}{\small GYAFC-EM}}
& Human Informal & 50.11 & 12.11 \\
~ & Human Formal & 53.76 & 15.82 \\
~ & ChatGPT Informal & 51.02$^\dag$$^\ddag$ & 12.01$^\ddag$ \\
~ & ChatGPT Formal & 51.98$^\dag$$^\ddag$ & 29.80$^\dag$$^\ddag$ \\
\bottomrule
\end{tabular}
\end{threeparttable}
}
\caption{Text formality scores by automatic metrics; $^\dag$ and $^\ddag$ indicate statistical significance (p$<$0.05) against same-style human answers, and opposite-style ChatGPT answers via paired t-test, respectively.}
\label{tab: Text Formality}
\end{table}

\subsubsection{Comparison to Previous SOTA Model}
We also find that ChatGPT outperforms the previous supervised SOTA model \cite{nouri-2022-text} by training on the same subset at Section \ref{4.4.1} for few-shot learning, as evident from the higher BLEU score. Specifically, ChatGPT yields superior scores of 0.711 and 0.697 in the EM and FR branches, as compared to the SOTA model's scores of 0.671 and 0.652. However, ChatGPT achieved only 0.07 and 0.06 BLEU scores on the EM and FR branches, respectively, in the unsupervised setting.

\subsubsection{Effect of Example Inclusion in Prompt}
We again examined the impact of including an example of the dataset into the prompt and find that this again helps ChatGPT slightly with matching the dataset style (with details provided in Table \ref{tab: Text Formality of One-shot Guidance}). Specifically, the formality score for the informal style is 50.67, while it climbs to 52.13 for the formal style, with the MTLD score also displaying an increase from 14.81 for informal texts to 19.22 for formal texts.

\subsubsection{Disparities in Style Transfer Behavior}

In terms of controlling the formality of sentence style, ChatGPT's performance still exhibits significant differences compared to human behavior. While the by-item correlation is slightly higher for this dataset than for the summary task (Pearson correlation of around 0.4 for formal style and 0.5 for informal style on the Formality metric; 0.3 for MTLD measure), there are interesting disparities between the distributions of POS tags between ChatGPT and humans. The examination of statistical significance further substantiates our antecedent observation, indicating a substantial disparity between the different response styles engendered by the model, as well as between the answers conforming to the same styles exhibited by humans.

\begin{figure}[htb]
  \centering  
  \includegraphics[width=0.5\textwidth,height=0.2\textwidth]{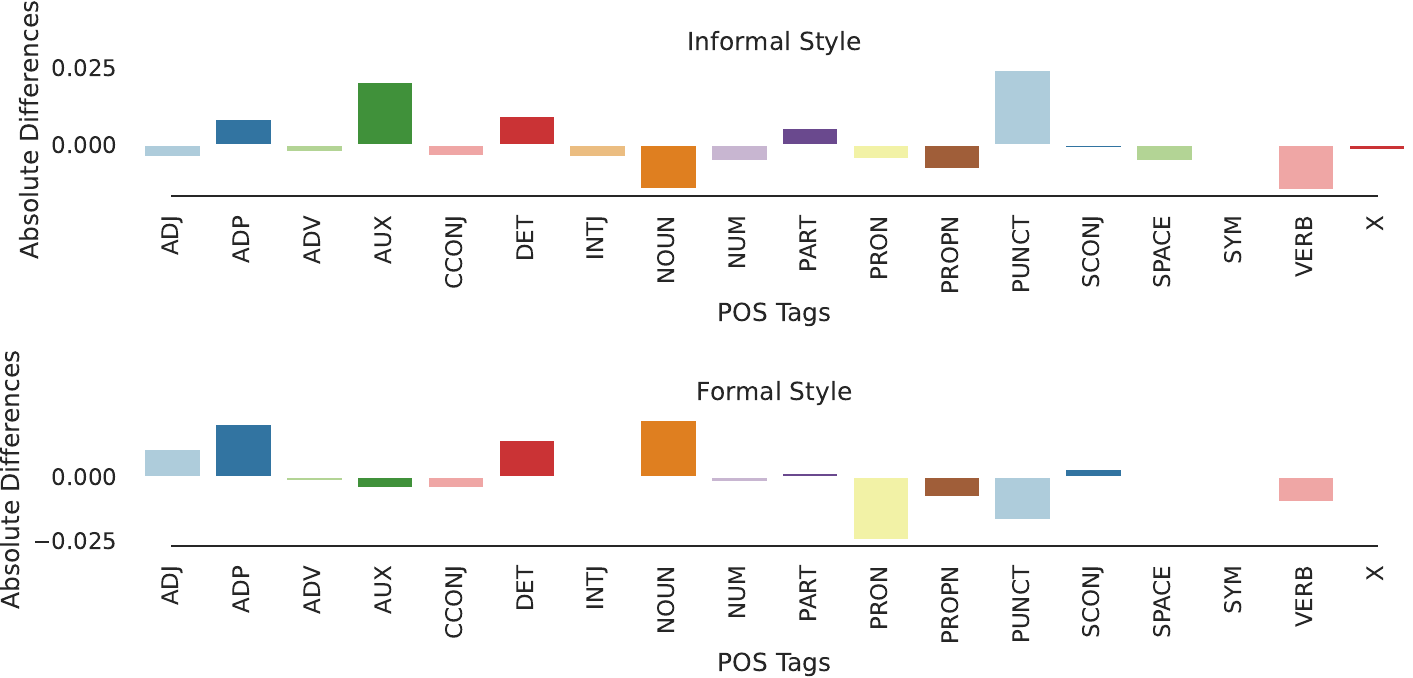}
  \caption{Absolute differences in POS tags distribution of ChatGPT and human-generated sentences: GYAFC - EM}
  \label{fig:EM_POS_formal}
\end{figure}

Figure \ref{fig:EM_POS_formal} illustrates the absolute differences in the distribution of Part-of-Speech (POS) tags. Based on this figure, it is evident that ChatGPT employs a higher frequency of adjectives, adpositions, determiners, and nouns in the generation of formal sentences when compared to those produced by human writers. Conversely, in the generation of informal sentences, ChatGPT tends to utilize more auxiliary words and punctuation marks. These variances in word choice between formal and informal styles, as exemplified by ChatGPT, are indicative of differences in its selected vocabulary for distinct stylistic modes compare with humans.

By analyzing the distribution of dependency labels (Appendix Figures \ref{fig:EM_DEP_formal}, \ref{fig:EM_DEP_informal}, \ref{fig:FR_DEP_formal}, \ref{fig:FR_DEP_informal}), it is also clear that, in comparison to human-authored sentences, ChatGPT utilizes a greater frequency of adjectival modifiers, auxiliaries, determiners, objects of the preposition, and prepositional modifiers for formal sentences. Contrarily, compounds and dependents are infrequently employed in the generation of informal sentences by ChatGPT.

\subsubsection{Inconsistencies and Hallucinations}

\begin{figure}[ht]
  \centering
  \includegraphics[width=0.5\textwidth]{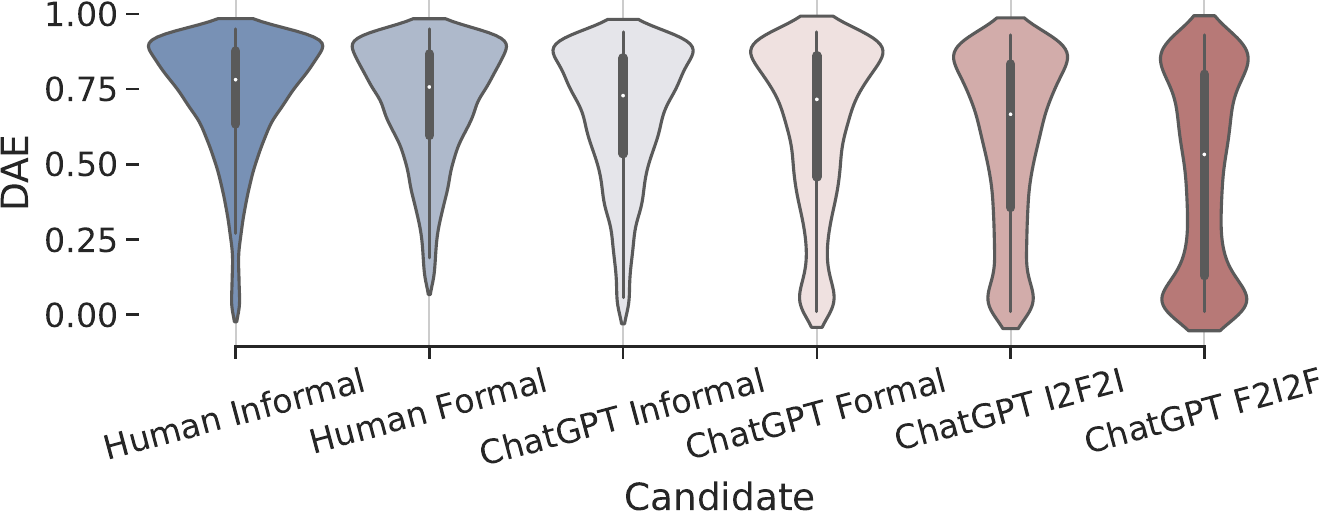}
  \caption{Dependency arc entailment: GYAFC - EM. Data points$>$0.95$\approx$Accurate. To clarify discrepancies, cutoff point$=$0.95.}
  \label{fig:Dependency_Arc_Entailment_EM}
\end{figure}

In order to assess the risk of introducing erroneous information when ChatGPT performs sentence style transformation, we employed DAE \cite{goyal-durrett-2020-evaluating} at the sentence level to examine the factuality after text style transformation, and compare again the effect of multiple re-writes.
Similar to before, F denotes formal style, I signifies informal style, and X2X2X (X $\in$ \{F, I\}) represents multiple rewriting transformations of the text. The outcomes of our inquiry are depicted in Figure \ref{fig:Dependency_Arc_Entailment_EM}, and Appendix Figure \ref{fig:Dependency_Arc_Entailment_FR}. We also again scrutinized the potential incorporation of hallucinatory information regarding named entities in the ChatGPT-generated text, and the findings are presented in Appendix Table \ref{tab: Named Entity Hallucination - GYAFC}.

Upon conducting factuality checking (see Figure \ref{fig:Dependency_Arc_Entailment_EM}, and Appendix Figure \ref{fig:Dependency_Arc_Entailment_FR}), it is discovered that ChatGPT's performance is inferior to that of humans in sentence-style rewriting. Interestingly, with the increase in the number of text conversions and rewritings, ChatGPT's tendency to commit factual errors escalates while the output increasingly deviates from the original text, compromising the fidelity of the final result. In a particular instance, the human-generated formal expression states ``She is a poor vocalist", whereas the formal rendition provided by ChatGPT articulates ``She does not possess the ability to sing". This discrepancy represents a significant semantic alteration. The degree of dependency arc entailment is low in this case. Similarly, Appendix Table \ref{tab: Named Entity Hallucination - GYAFC} reveals that recall scores on the named entity hallucination metric are lower in ChatGPT sentences than in human sentences.

\subsubsection{Qualitative Examples}
To explore whether ChatGPT-generated sentences significantly alter the original semantics of the input text, we conducted a case study by randomly selecting 15 samples from each branch of the GYAFC dataset. Our findings indicate that ChatGPT poses a relatively severe risk of modifying the original semantics during sentence style transformation, with approximately 18\% of the samples exhibiting noticeable semantic inconsistencies. The examples in Table \ref{tab:case_study} reveal that during the process of sentence style transfer, ChatGPT erroneously modifies the content words, resulting in significant semantic alterations. 

\begin{table}[htbp]
\centering
\scriptsize
\scalebox{1.1}{
\begin{tabular}{lr}
\toprule
Formal to Informal\\
 \midrule
  It is such a waste of TV space. &(Reference) \\
  Yes, because it's such a waste of TV space!& (Human) \\
  What a total waste of TV \textcolor{red}{time}!& (ChatGPT) \\
 \midrule
 The other boy isn't that great.& (Reference) \\
 The other boy is not that good.& (Human) \\
 The other kid's not so \textcolor{red}{hot}.& (ChatGPT) \\
\midrule
 I really enjoy how the composition has the tec...& (Reference) \\
 I really like how they do like the whole techn...& (Human) \\
 I'm \textcolor{red}{diggin' how the techno beat slows down in} ...&(ChatGPT) \\
\midrule
\midrule
{Informal to Formal} \\\hline
 Fatboy Slim - Right Here, Right Now Or any oth... &(Reference) \\
 Fatboy Slim is right here and now. He Rocks!& (Human) \\
 Fatboy Slim\textcolor{red}{'s "Right Here, Right Now"} is an ex... &(ChatGPT) \\
\midrule
 loved them since their first album. &(Reference) \\
 I have loved them since their first album.& (Human) \\
 I have held a fondness for them since the \textcolor{red}{rele...}&(ChatGPT) \\
\midrule
 if u occasionally doing it then u alrady r add... &(Reference) \\
 If you occasionally do it, then you are already... &(Human) \\
 If you are \textcolor{red}{engaging in the activity} on a regul...&(ChatGPT) \\
\bottomrule
\end{tabular}
}
\caption{\label{tab:case_study} Case study of ChatGPT generated output}
\end{table}

Furthermore, our examination of the visualized dependency tree (see Appendix Figures \ref{fig:cs1_ref}, \ref{fig:cs1_hum}, and \ref{fig:cs1_chat}), which relies primarily on the dependency arc entailment (DAE) algorithm for fact-checking, reveals that the text generated by ChatGPT contains a higher number of dependency arcs lacking support from the original text, when compared to human responses. 

\section{Conclusion}

This paper presents a broad assessment of ChatGPT's proficiency in generating controllable text. We conducted quantitative and qualitative examinations at the document level (summarization task) and sentence level (text style transfer). The empirical findings show that ChatGPT outperforms the previous state-of-the-art models on automatic metrics, but that there are substantial disparities between its generated texts and human-written texts. These disparities are reduced by providing a target example of the human writing style. Furthermore, our investigations also confirm the previously reported problems of hallucinations and inaccuracies in text generated by ChatGPT.

\section{Limitations}

The primary limitations of the current study pertain to the selection of prompts and evaluation metrics. The experimental cost of requesting API responses from OpenAI to assess ChatGPT's text generation abilities imposes significant constraints on our choice of datasets. Therefore, we have to limit our experimentation to only two related controllable text generation datasets. While we have evaluated ChatGPT's performance at both the document and sentence levels, we cannot extrapolate that ChatGPT has similar performance for other text generation datasets. Additionally, the experimental cost prohibits us from conducting traversal experiments on the selection of hyperparameters. We relied on the default configuration recommended by OpenAI, and we maintain consistency in all hyperparameters to ensure the fairness of the experiments.

Secondly, although we have studied the impact of prompt engineering on ChatGPT, the selection of prompts is mainly affected by human understanding, and the number of potential prompts is infinite. Hence, we cannot guarantee whether other prompts that we did not select will yield the same conclusions as our experiment. Furthermore, ChatGPT is subject to continuous updates and iterations, which may lead to improved performance, making it difficult to predict if future versions of ChatGPT will have similar results to our experiments.

Finally, to select appropriate evaluation metrics, we have included both domain-related evaluation metrics (such as reading difficulty and text formality) and domain-independent evaluation indicators (such as fact-checking and hallucination detection). However, we acknowledge that the automatic metrics may sometimes not capture all aspects of the intended construct correctly.

\section{Ethics Considerations}

All datasets utilized in this study are publicly available, and we have adhered to ethical considerations by not introducing any additional information into ChatGPT's inputs.

\section*{Acknowledgements} 
This project has received funding from the European Research Council (ERC) under the European Union’s Horizon 2020 Research and Innovation Programme (Grant Agreement No. 948878).
\begin{figure}[H] 
\centering
\includegraphics[width=0.5\columnwidth]{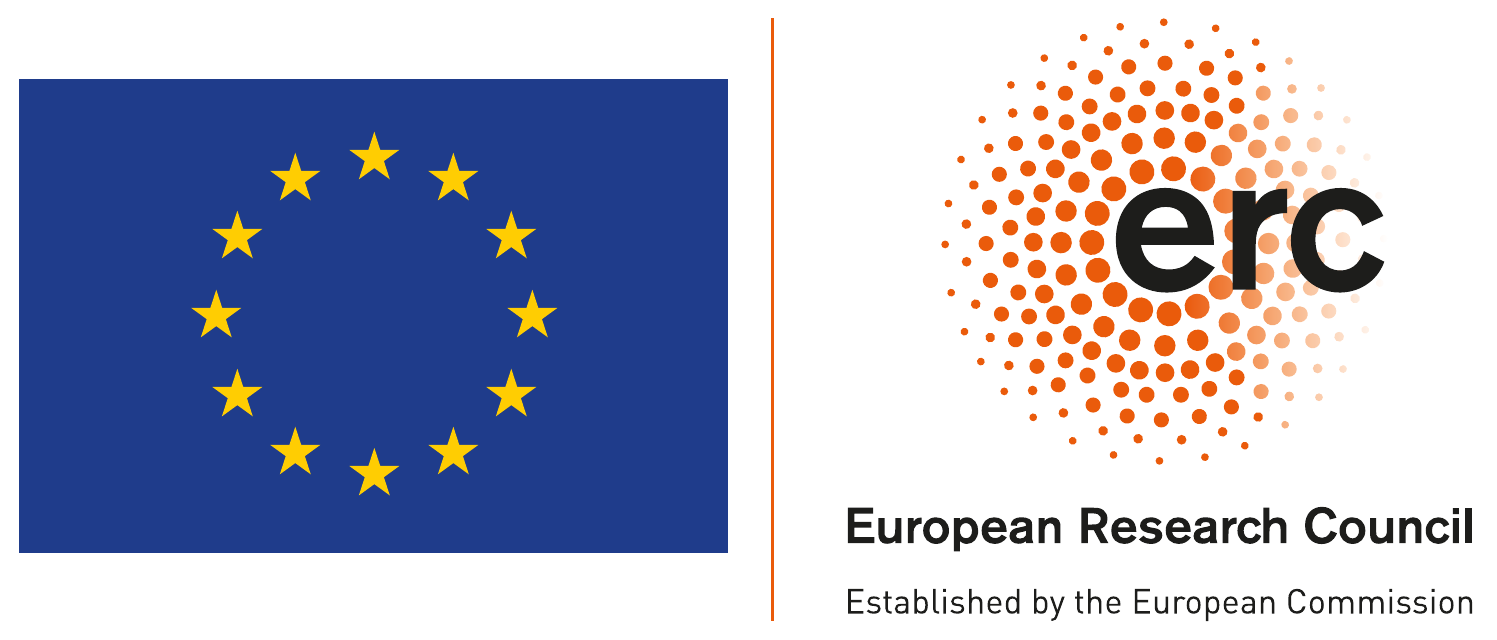}
\end{figure}

\bibliography{anthology,custom}
\bibliographystyle{acl_natbib}

\appendix

\section{Appendix: One-shot Guidance}
\label{sec:appendixB}

\begin{table*}[htbp]
\centering
\scalebox{0.78}{\tabcolsep=4pt
\begin{threeparttable}
\begin{tabular}{l c c c}
\toprule
Candidate & FRE & CLI & DCR \\
\hline
Document: \{Original Document\}, Layman Summary: \{Human Layman Summary\}.\\ Please learn the way of summarization from the previous example, and give me a layman-style summary of X & 49.23$^\dag$ & 13.26$^\dag$ & 10.45$^\dag$\\
Human Answer & 52.42 & 12.46 & 8.93\\
\hdashline
Document: \{Original Document\}, Expert Summary: \{Human Expert Summary\}.\\ Please learn the way of summarization from the previous example, and give me an expert-style summary of X & 28.88$^\dag$& 15.92$^\dag$ & 11.82\\
Human Answer & 23.20 & 17.62 & 11.78 \\
\bottomrule
\end{tabular}
\end{threeparttable}
}
\caption{Reading difficulty of one-shot guidance. $^\dag$ indicates statistical significance (p$<$0.05) against corresponding human answers via paired t-test.}
\label{tab:Reading Difficulty of One-shot Guidance}
\end{table*}

\begin{table*}[htbp]
\centering
\scalebox{0.8}{\tabcolsep=4pt
\begin{threeparttable}
\begin{tabular}{l c c}
\toprule
Candidate & Formality & MTLD \\
\hline
Formal: \{Formal Sentence\}, Informal: \{Informal Sentence\}.\\ Please learn the way of formality conversion from the previous example, and give me an informal version of X & 50.67$^\dag$ & 14.81 \\
Human Answer & 49.87 & 15.20 \\
\hdashline
Informal: \{Informal Sentence\}, Formal: \{Formal Sentence\}.\\ Please learn the way of formality conversion from the previous example, and give me a formal version of X & 52.13$^\dag$ & 19.22 \\
Human Answer & 53.57 & 18.70 \\
\bottomrule
\end{tabular}
\end{threeparttable}
}
\caption{Text formality of one-shot guidance on GYAFC-FR branch. $^\dag$ indicates statistical significance (p$<$0.05) against corresponding human answers via paired t-test.}
\label{tab: Text Formality of One-shot Guidance}
\end{table*}

\section{Appendix: Absolute Differences in POS and Dependency Label Distributions}
\label{sec:appendixD}

\FloatBarrier
\begin{figure*}[htbp]
  \centering
  \includegraphics[width=1\textwidth,height=0.4\textwidth]{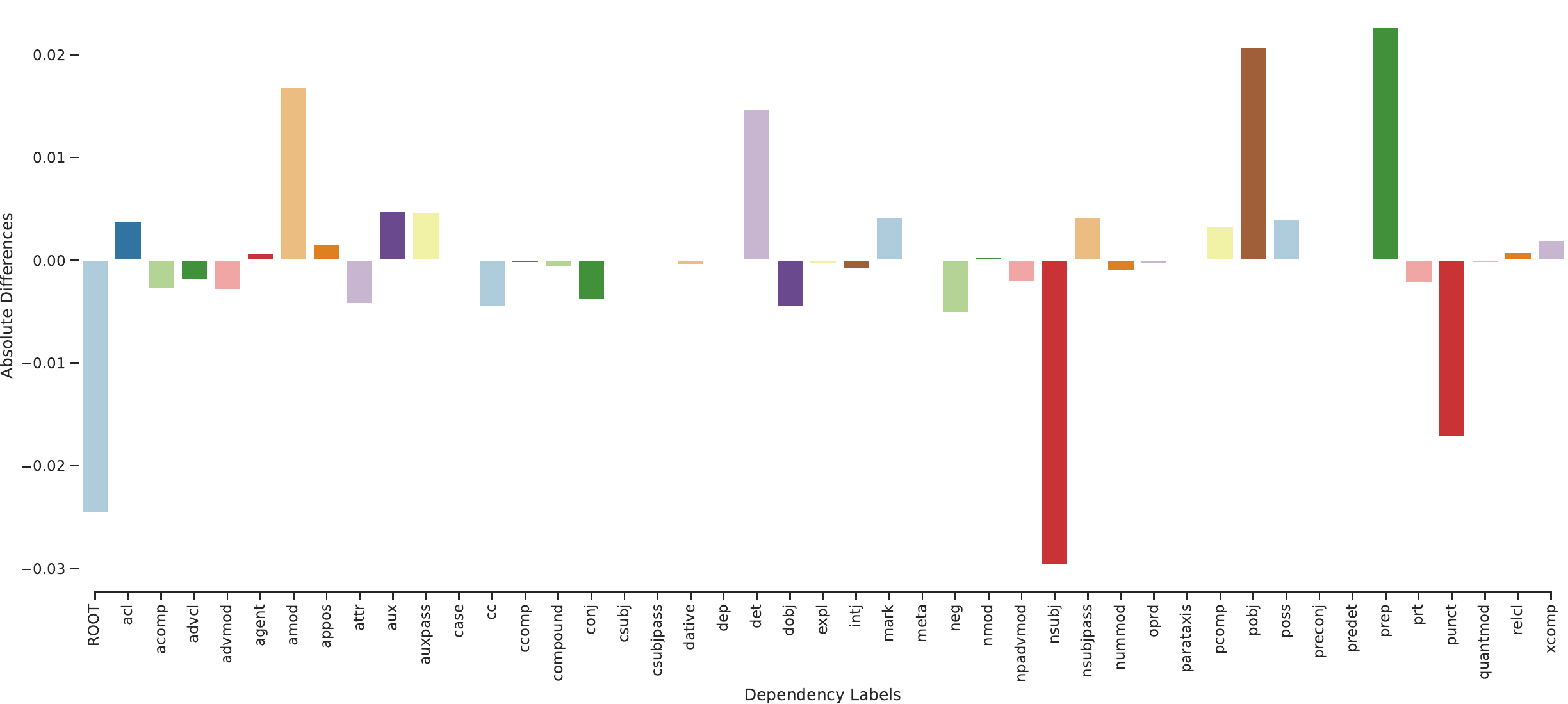}
  \caption{Absolute differences in dependency labels distribution of ChatGPT and human-generated formal style sentences: GYAFC - EM }
  \label{fig:EM_DEP_formal}
\end{figure*}

\begin{figure*}[htbp]
  \centering
  \includegraphics[width=1\textwidth,height=0.45\textwidth]{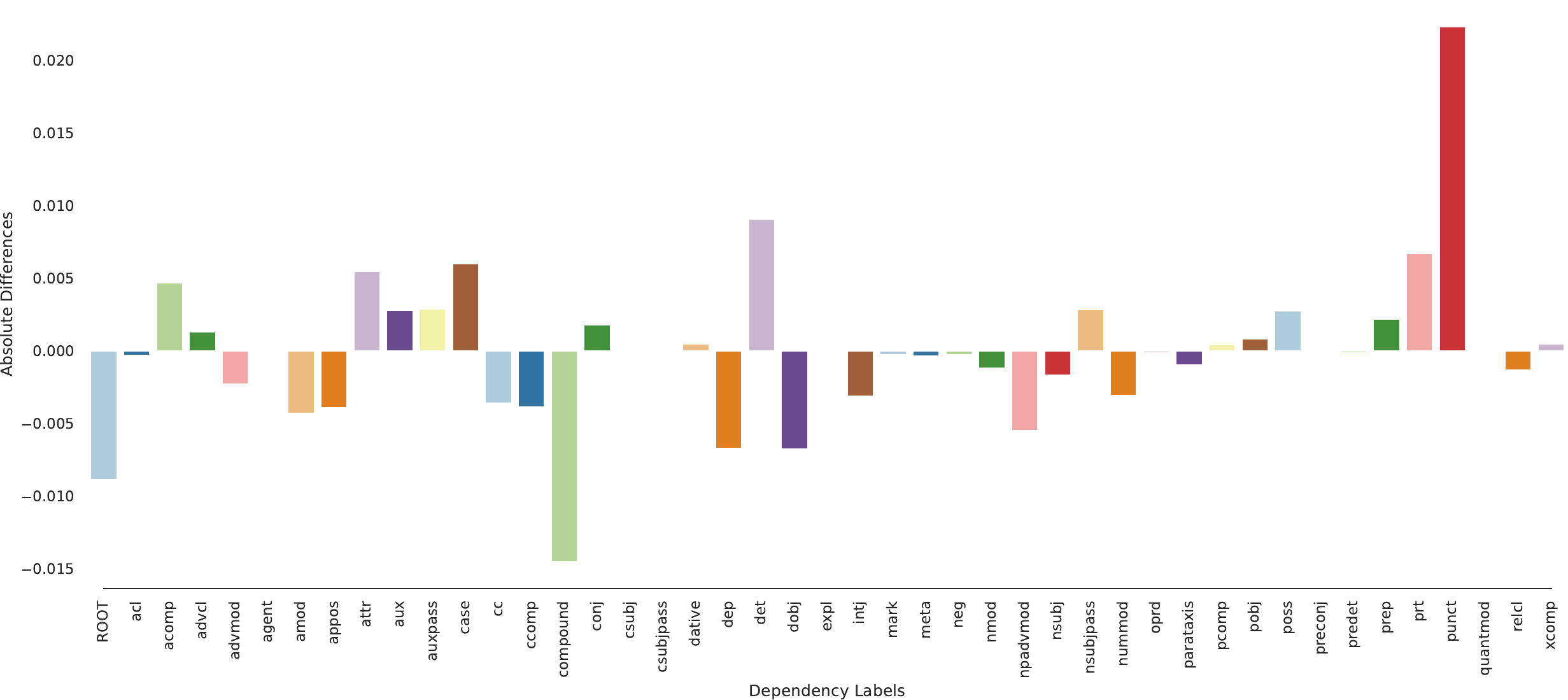}
  \caption{Absolute differences in dependency labels distribution of ChatGPT and human-generated informal style sentences: GYAFC - EM}
  \label{fig:EM_DEP_informal}
\end{figure*}

\begin{figure*}[htbp]
  \centering
  \includegraphics[width=1\textwidth,height=0.4\textwidth]{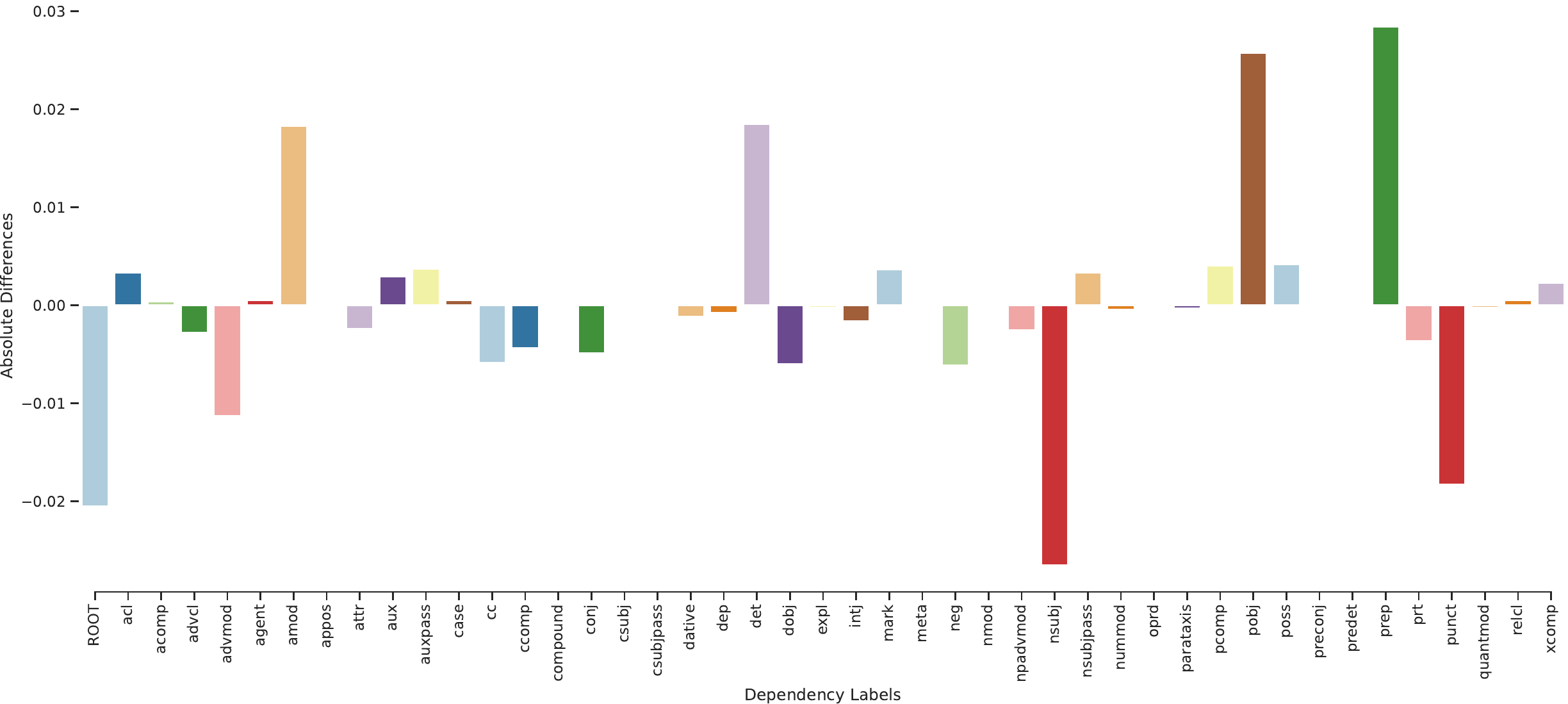}
  \caption{Absolute differences in dependency labels distribution of ChatGPT and human-generated formal sentences: GYAFC - FR}
  \label{fig:FR_DEP_formal}
\end{figure*}

\begin{figure*}[htbp]
  \centering
  \includegraphics[width=1\textwidth,height=0.4\textwidth]{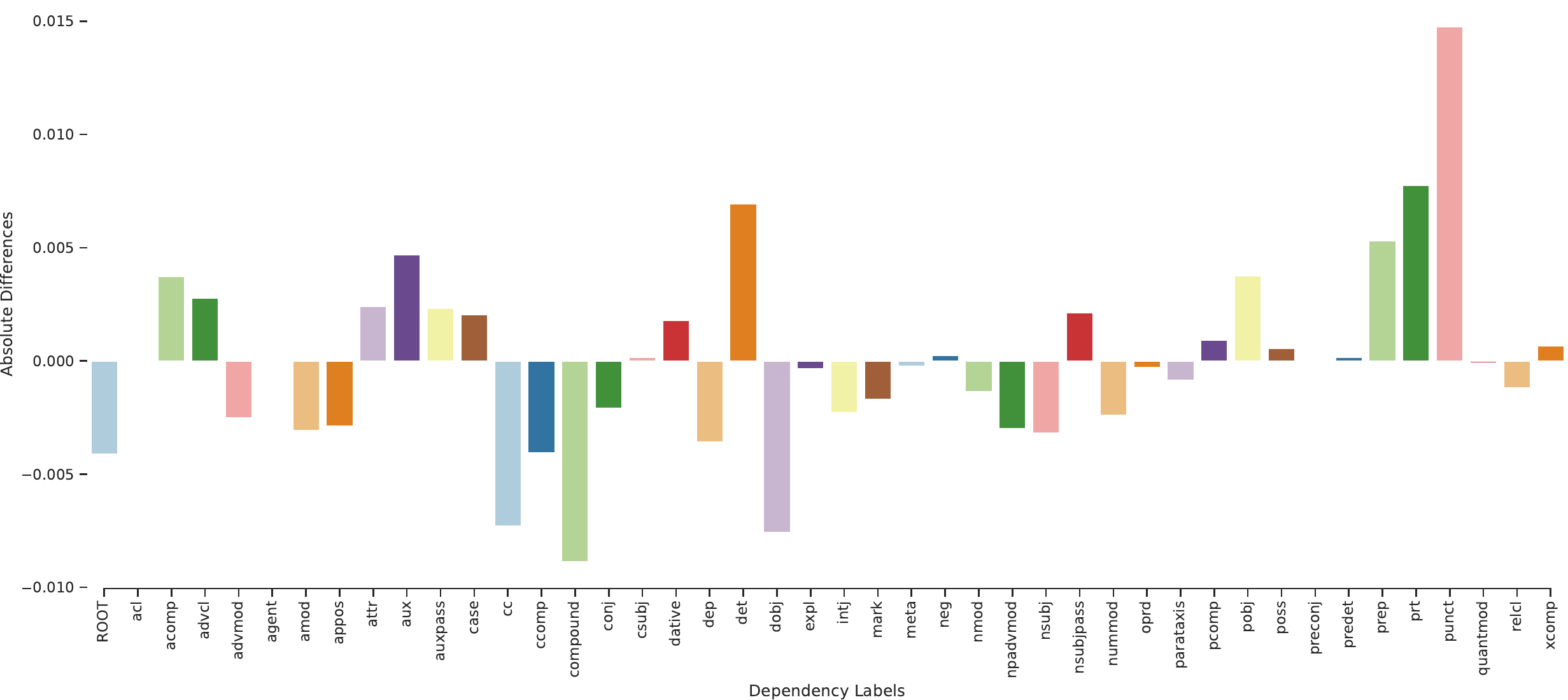}
  \caption{Absolute differences in dependency labels distribution of ChatGPT and human-generated informal sentences: GYAFC - FR}
  \label{fig:FR_DEP_informal}
\end{figure*}

\FloatBarrier
\begin{figure}[htbp]
  \centering
  \includegraphics[width=0.5\textwidth,height=0.33\textwidth]{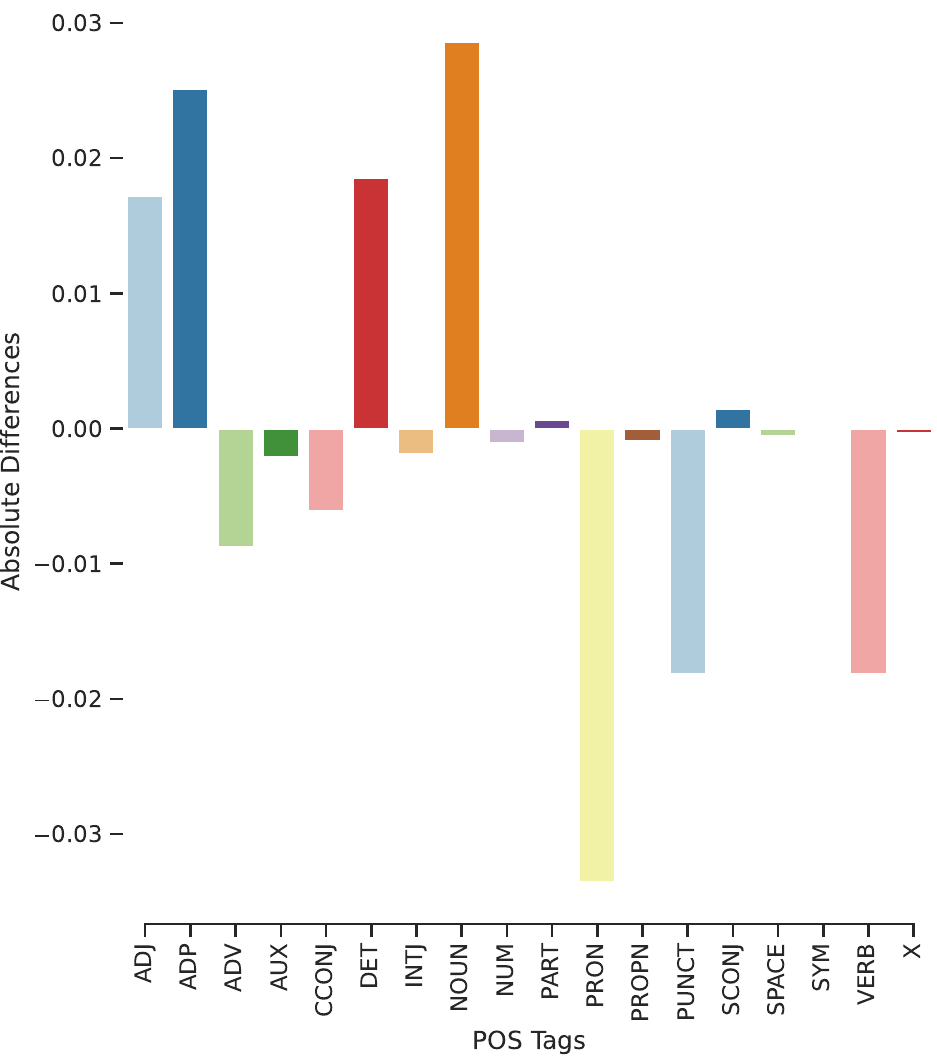}
  \caption{Absolute differences in POS tags distribution of ChatGPT and human-generated formal sentences: GYAFC - FR}
  \label{fig:FR_POS_formal}
\end{figure}

\begin{figure}[htbp]
  \centering
  \includegraphics[width=0.5\textwidth,height=0.33\textwidth]{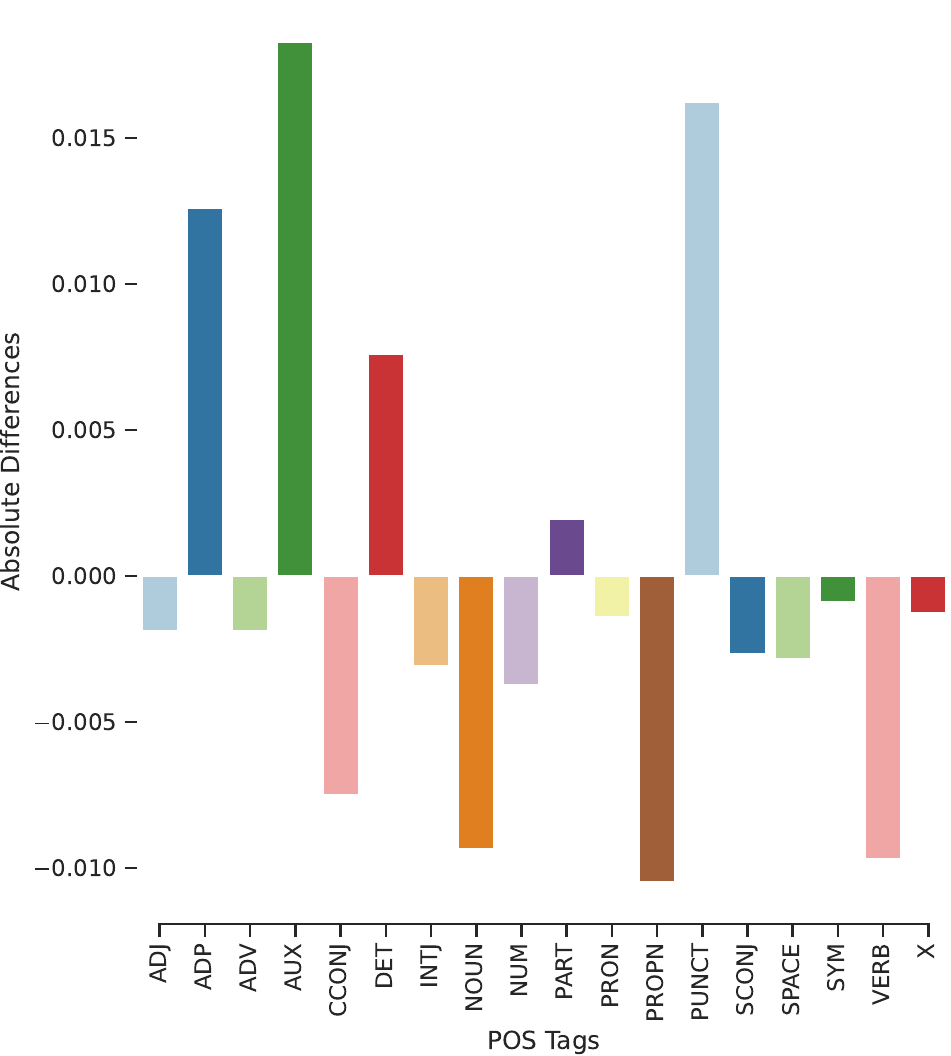}
  \caption{Absolute differences in POS tags distribution of ChatGPT and human-generated informal sentences: GYAFC - FR}
  \label{fig:FR_POS_informal}
\end{figure}

\begin{figure*}[htbp]
  \centering
  \includegraphics[width=1\textwidth,height=0.45\textwidth]{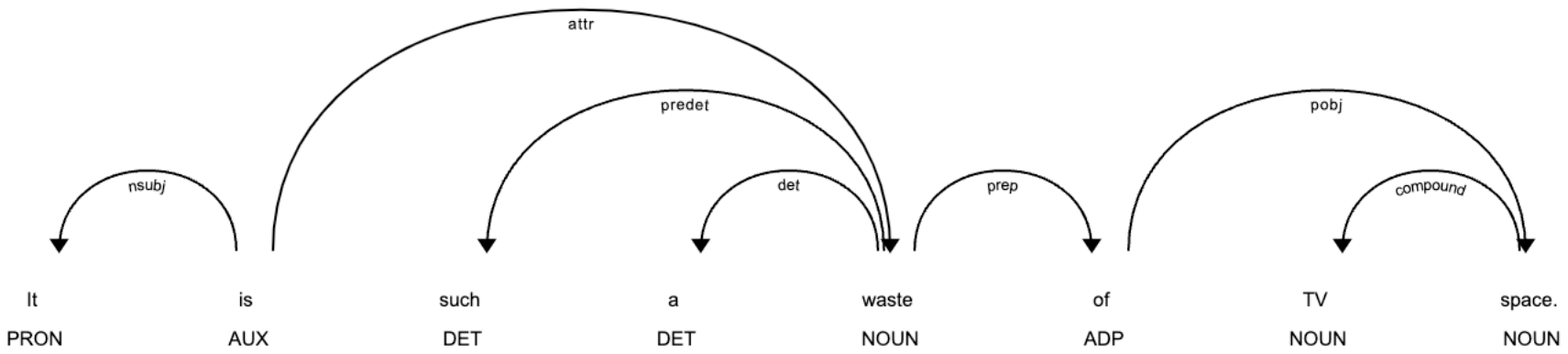}
  \caption{Case study of dependency tree visualization (Reference)}
  \label{fig:cs1_ref}
\end{figure*}

\begin{figure*}[htbp]
  \centering
  \includegraphics[width=1\textwidth,height=0.45\textwidth]{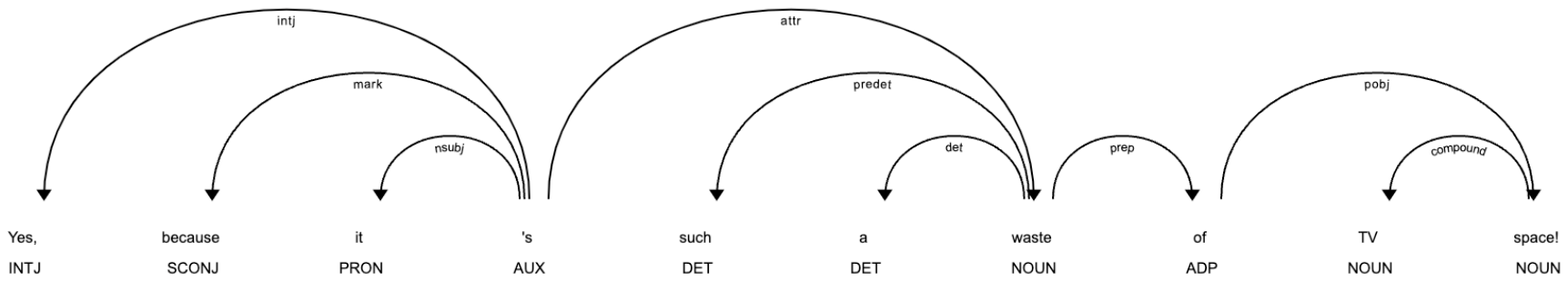}
  \caption{Case study of dependency tree visualization (Human)}
  \label{fig:cs1_hum}
\end{figure*}

\begin{figure*}[htbp]
  \centering
  \includegraphics[width=1\textwidth,height=0.45\textwidth]{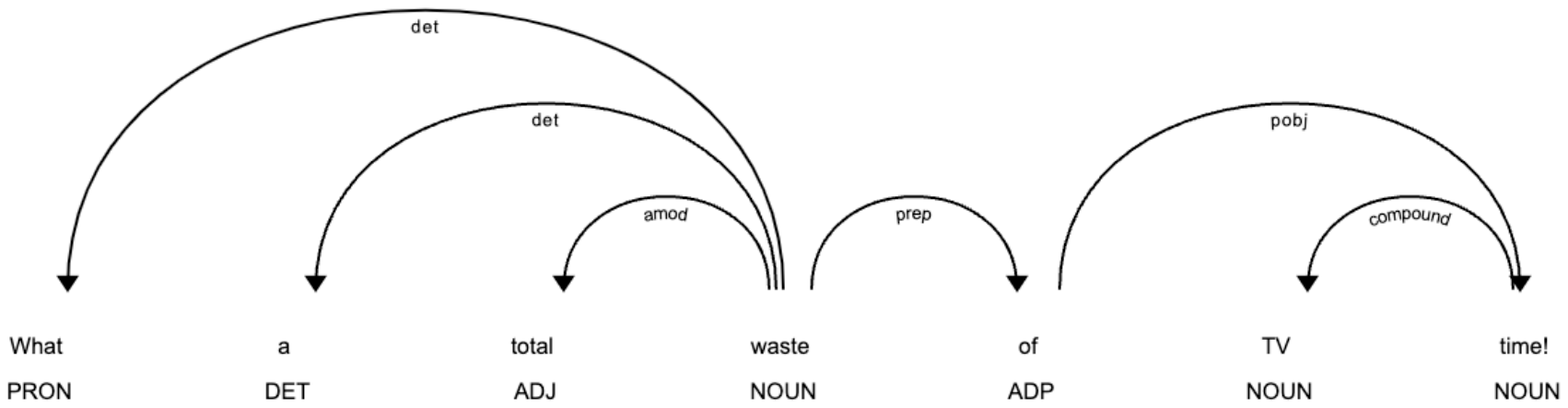}
  \caption{Case study of dependency tree visualization (ChatGPT)}
  \label{fig:cs1_chat}
\end{figure*}

\section{Appendix: Dependency Arc Entailment}
\label{sec:appendixF}
\FloatBarrier
\begin{figure}[H]
  \centering
  \includegraphics[width=0.5\textwidth,height=0.3\textwidth]{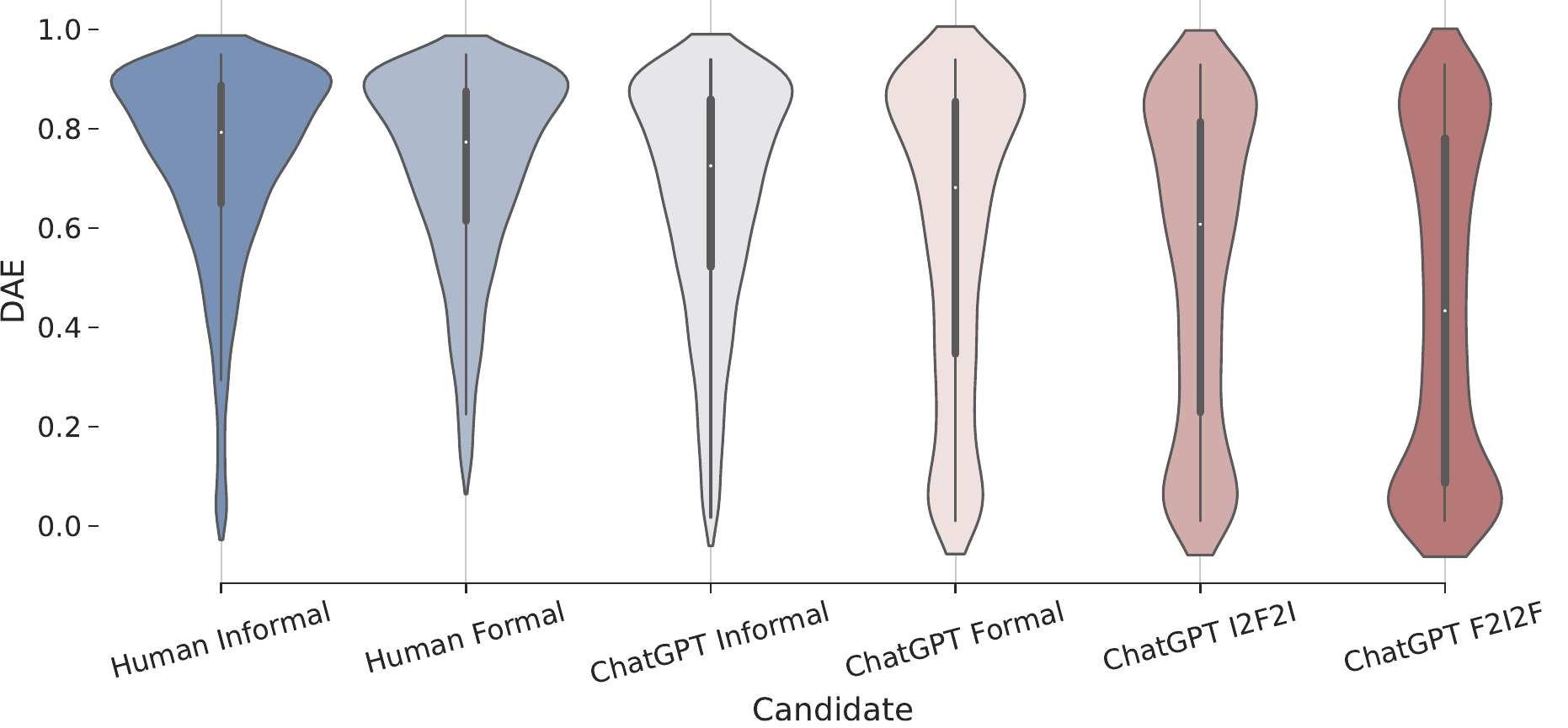}
  \caption{Dependency arc entailment: GYAFC - FR. Data points$>$0.95$\approx$Accurate. To clarify discrepancies, cutoff point$=$0.95.}
  \label{fig:Dependency_Arc_Entailment_FR}
\end{figure}

\section{Appendix: Named Entity Hallucination}
\label{sec:appendixG}
\FloatBarrier
\begin{table}[htbp]
\centering
\scalebox{0.85}{\tabcolsep=4pt
\begin{threeparttable}
\begin{tabular}{c c c c c}
\toprule
Dataset & Candidate & Precision & Recall & F1 \\
\hline
\multirow{6}*{\rotatebox{90}{GYAFC-FR}}
& Human Informal & 0.989 & 0.988 & 0.988 \\
~ & Human Formal & 0.988 & 0.989 & 0.988 \\
~ & ChatGPT Informal & 0.986 & 0.985 & 0.986 \\
~ & ChatGPT Formal & 0.974 & 0.974 & 0.974 \\
~ & ChatGPT I2F2I & 0.982 & 0.982 & 0.982 \\
~ & ChatGPT F2I2F & 0.974 & 0.973 & 0.973 \\
\midrule
\midrule
\multirow{6}*{\rotatebox{90}{GYAFC-EM}}
& Human Informal & 0.979 & 0.987 & 0.983 \\
~ & Human Formal & 0.977 & 0.989 & 0.982 \\
~ & ChatGPT Informal & 0.975 & 0.974 & 0.974 \\
~ & ChatGPT Formal & 0.950 & 0.952 & 0.951 \\
~ & ChatGPT I2F2I & 0.970 & 0.969 & 0.970 \\
~ & ChatGPT F2I2F & 0.945 & 0.946 & 0.945 \\
\bottomrule
\end{tabular}
\end{threeparttable}
}
\caption{Named entity hallucination - GYAFC}
\label{tab: Named Entity Hallucination - GYAFC}
\end{table}

\end{document}